\documentclass[11pt,a4paper]{article}

\usepackage[hyperref]{naaclhlt2018}

\usepackage{times}
\usepackage{latexsym}
\usepackage{amsmath}

\usepackage{url}
\usepackage{graphicx}

\usepackage{xspace}
\newcommand*{\eg}{e.g.\@\xspace}

\makeatletter
\newcommand*{\etc}{%
	\@ifnextchar{.}%
	{etc}%
	{etc.\@\xspace}%
}
\makeatother

\usepackage{booktabs}
\usepackage[capitalise]{cleveref}

\usepackage{tikz}
\usepackage{tabularx}

\newcommand{\mslp}{\textsc{slp}}
\newcommand{\mflip}{\textsc{flip}}
\newcommand{\mortho}{\textsc{ortho}}
\newcommand{\mconv}{\textsc{conv}}
\newcommand{\mproj}{\textsc{proj}}

\aclfinalcopy

\title{Extrapolation in NLP}

\author{Jeff Mitchell, Pasquale Minervini, Pontus Stenetorp \and Sebastian Riedel \\
        University College London \\ Department of Computer Science \\ 
        \tt \{j.mitchell, p.minervini, p.stenetorp, s.riedel\}@cs.ucl.ac.uk}

\date{}

\begin{document}
\maketitle
\begin{abstract}
We argue that extrapolation to examples outside the training space will often be easier for models that capture global structures, rather than just maximise their local fit to the training data.
We show that this is true for two popular models: the Decomposable Attention Model and word2vec.
\end{abstract}

\section{Introduction}

%
%
%
%

%
%
%

%
In a controversial essay, \citet{marcus2017} draws the distinction between two types of generalisation: \emph{interpolation} and \emph{extrapolation}; with the former being predictions made \emph{between} the training data points, and the latter being generalisation \emph{outside} this space.
He goes on to claim that deep learning 
is only effective at interpolation, but that human like learning and behaviour requires extrapolation.
%
%

%
On Twitter, Thomas Diettrich rebutted this claim with the response that no methods extrapolate; that \emph{what appears to be extrapolation from X to Y is interpolation in a representation that makes X and Y look the same.}~\footnote{\href{https://twitter.com/tdietterich/status/948811920001282049}{https://twitter.com/tdietterich/\\status/948811920001282049}}
It is certainly true that extrapolation is hard, but there appear to be clear real-world examples.
For example, in 1705, using Newton's then new inverse square law of gravity, Halley predicted the return of a comet 75 years in the future.
This prediction was not only possible for a new celestial object for which only a limited amount of data was available, but was also effective on an orbital period twice as long as any of those known to Newton.
%
%
Pre-Newtonian models required a set of parameters (deferents, epicycles, equants, \etc) for each body 
and so would struggle to generalise from known objects to new ones.
Newton's theory of gravity, in contrast, not only described celestial orbits but also predicted the motion of bodies thrown or dropped on Earth.

In fact, most scientists would regard this sort of extrapolation to new phenomena as a vital test of any theory's legitimacy.
Thus, the question of what is required for extrapolation is reasonably important for the development of NLP and deep learning.

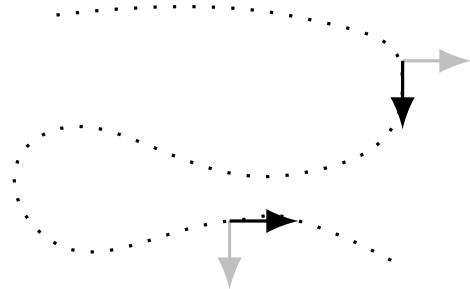
\begin{figure}[t]
	\centering
	\resizebox{0.9\columnwidth}{!}{
	\begin{tikzpicture}
	  \draw [dotted] plot [smooth, tension=1] coordinates {(0,-0.5) (1,-0.5) (1.5,-0.8)  (1,-1.2) (0,-1) (0,-1.5) (0.75,-1.4) (1.05,-1.4) (1.5,-1.6)};
	  \draw [lightgray,-latex] (1.5,-0.7) -- (1.8,-0.7);
	  \draw [-latex] (1.5,-0.7) -- (1.5,-1.0);
	  \draw [lightgray,-latex] (0.75,-1.4) -- (0.75,-1.7);
	  \draw [-latex] (0.75,-1.4) -- (1.05,-1.4);
	\end{tikzpicture}
	}\caption{Generalising to unseen data: dotted line = training manifold; black arrows = interpolation; grey arrows = extrapolation. Both directions are represented globally in the training data, but local interpolation is only effective in one of them at each point.
	}\label{extrafig}
\end{figure}
\citet{marcus2017} proposes an experiment, consisting of learning the identity function for binary numbers, where the training set contains only the even integers but at test time the model is required to generalise to odd numbers.
A standard multilayer perceptron (MLP) applied to this data fails to learn anything about the least significant bit in input and output, as it is constant throughout the training set,
and therefore fails to generalise to the test set.
Many readers of the article ridiculed the task and questioned its relevance.
Here, we will argue that it is surprisingly easy to solve Marcus' even-odd task and that the problem it illustrates is actually endemic throughout machine learning.
\citet{marcus2017} links his experiment to the systematic ways in which the meaning and use of a word in one context is related to its meaning and use in another~\citep{fodorpylyshyn1988,lakebaroni2017}.
These regularities allow us to extrapolate from 
sometimes even a single use of a word to understand all of its other uses. 
In fact, we can often use a symbol effectively with no prior data.
For example, a language user that has never have encountered the symbol \emph{Socrates} before may nonetheless be able to leverage their syntactic, semantic and inferential skills to conclude that \emph{Socrates is mortal} contradicts \emph{Socrates is not mortal}.

Marcus' experiment essentially requires extrapolating what has been learned about one set of symbols to a new symbol in a systematic way.
However, this transfer is not facilitated by the techniques usually associated with improving generalisation, such as L2-regularisation \cite{l2reg1963}, drop-out \cite{dropout2014} or preferring flatter optima \cite{flatopt1995}.

In the next section, we present four ways to solve this problem and discuss the role of global symmetry in effective extrapolation to the unseen digit. 
Following that we present practical examples of global structure in the representation of sentences and words.
Global, in these examples, means a model form that introduces dependencies between distant regions of the input space.

\section{Four Ways to Learn the Identity Function}

The problem is described concretely by \citet{marcus1998}, 
with inputs and outputs both consisting of five units representing the binary digits of the integers zero to thirty one.
The training data consists of the binary digits of the even numbers $(0, 2, 4, 8, \ldots, 30)$ and the test set consists of the odd numbers $(1, 3, 5, 7, \ldots, 31)$.
The task is to learn the identity function from the training data in a way that generalises to the test set.
The first model (\mslp) we consider is a simple linear single layer perceptron from input to output.
In the second model (\mflip), we employ a change of representation.
Although the inputs and outputs are given and fixed in terms of the binary digits \textbf{1} and \textbf{0}, we will treat these as symbols and exploit the freedom to encode these into numeric values in the most effective way for the task.
Specifically, we will represent the digit \textbf{1} with the number \texttt{0} and the digit \textbf{0} with the number \texttt{1}.
Again, the network will be a linear single layer perceptron without biases. 
Returning to the original common-sense representation, \textbf{1} $\rightarrow$ \texttt{1} and \textbf{0} $\rightarrow$ \texttt{0}, the third model (\mortho) attempts to improve generalisation by imposing a global condition on the matrix of weights in the linear weights.
In particular, we require that the matrix is orthogonal, and apply the absolute value function at the output to ensure the outputs are not negative.
For the fourth model (\mconv), we use a linear Convolutional Neural Network (ConvNet, \citealp{Lecun98gradient-basedlearning}) with a filter of width five.
In other words, the network weights define a single linear function 
that is shifted across the inputs for each output position.
Finally, in our fifth model (\mproj) we employ another change of representation, this time a dimensionality reduction technique.
Specifically, we project the 5-dimensional binary digits $\mathbf{d}$ onto an $n$ dimensional vector $\mathbf{r}$ and carry out the learning using an $n$-to-$n$ layer in this smaller space.
\begin{equation} \label{dimred}
 \mathbf{r} = \mathbf{A} \mathbf{d}
\end{equation}
\noindent where the entries of the matrix $\mathbf{A}$ are $A_{ij} = e^{\beta (j - i)}$.
In each case, our loss and test evaluation is based on squared error between target and predicted outputs.
\begin{table}[t]
\centering
{%
 \begin{tabular}{lcc}
 \toprule
  {\bf Model} & {\bf Train} & {\bf Test} \\
 \midrule
  \mslp           & 8.12e-06 & 0.99 \\
  \mflip          & 6.79e-05 & 1.04e-05 \\
  \mortho         & 1.27e-04 & 4.09e-05 \\
  \mconv          & 1.71e-04 & 3.20e-05 \\
  \mproj          & 5.15e-06 & 8.07e-06 \\
 \bottomrule
 \end{tabular}
}%
\caption{Mean Squared Error on the Train (even numbers) and Test (odd numbers) Sets.}\label{msetab}
\end{table}

\paragraph{Training.} Each model is implemented in TensorFlow~\citep{tensorflow2015-whitepaper} and optimised for 1,000 epochs.
In \cref{dimred}, we find that values of $\beta=\ln(2)$ and $n=1$ work well in practice.
\paragraph{Results.}
As can be seen in \cref{msetab}, {\mslp} fails to learn a function that generalises to the test set.
In contrast, all the other models (\mflip, \mortho, \mconv, \mproj) generalise almost perfectly to the test set.
Thus, we are left with four potential approaches to learning the identity function.
Is lowest test set error the most appropriate means of choosing between them?
\paragraph{Discussion.}
This decision probably isn't as momentous as the choice discussed by Galileo in his Dialogue Concerning the Two Chief World Systems, where he presented the arguments for and against the heliocentric and geocentric models of planetary motion. 
%
%
These pre-Newtonian models could, in principle, attain as much predictive accuracy as desired, 
given enough data, by simply incorporating more epicycloids for each planet. 
On the other hand, they could not extrapolate beyond the bodies in that training data.
Here, we will try to extract something useful from our results by considering how each model might generalise to other data and problems.
%
%

%
%
%

%
%
%

%
%
Although {\mflip} has the second lowest test set error, it is at best a cheap hack\footnote{Nonetheless, such tricks are hardly unknown in machine learning research.} which works only in the limited circumstance of this particular problem.
If there were more than a single fixed digit in the training data, this trick would not work.

{\mortho} suffers from the same problem, 
though it does embody the principle that everything in the input should end up in the output which seems to be part of this task.

{\mconv} on the other hand will generalise to any size of input and output, and will even generalise to multiplication by powers of 2, rather than just learning the identity function.

{\mproj}, with the values $\beta=\ln(2)$ and $n=1$, boils down to converting the binary digits into the equivalent single real value and learning the identity function via linear regression.
This approach will extrapolate to values of any magnitude\footnote{Generalisation to values outside the training set would not be so successful had we used an MLP rather than a uniform linear function. Fitting to the training set using sigmoids will not yield a function that continues to approximate the identity very far beyond its range in the training set.} and generalise to learning any linear function, rather than just the identity.
As such, it is probably the only practically sensible solution, although it cheats by avoiding the central difficulty in the original problem.
At its most general, this central difficulty is the problem of extrapolating in a direction that is perpendicular to the training manifold.
The even number inputs lay on a 4 dimensional subspace,  while the odd numbers were displaced in a direction at right angles to that subspace.
In this general form, the problem of how to respond to variation in the test set that is perpendicular to the training manifold lacks a well-defined unique solution, and this helps to explain why many people dismissed the task entirely.
However, this problem is in fact pervasive in most of machine learning.
Training instances will typically lie on a low dimensional manifold and effective generalisation to new data sources will commonly require handling variation that is orthogonal to that manifold in an appropriate manner, e.g. \cref{extrafig}.
If prediction is based on local interpolation using a highly non-linear function, then no amount of smoothing of the fit will help.
Convolution is able to extrapolate from even to odd numbers because it exploits the key structure of the ordering of digits that a human would use.
A human, given this task, would recognise the correspondence between input and output positions and then apply the same copying operation at each digit, which is essentially what convolution learns to do.
It implicitly assumes that there is a global translational symmetry\footnote{Coincidentally, the rejection of the Earth centred model in favour of planetary motions orbiting the Sun played an important role in the recognition that the laws of physics also have a global translational symmetry, i.e. that no point in space is privileged or special.} across input positions, and this reduces the number of parameters and allows generalisation from one digit to another.
Returning to the linguistic question that inspired the task, 
we can think of systematicity in terms 
of symmetries that preserve the meaning of a word or sentence \cite{kiddondomingos2015}.
Ideally, our NLP models should embody or learn the symmetries that allow the same meaning to be expressed within multiple grammatical structures.

Unfortunately, syntax is complex and prohibits a short and clear investigation here.
On the other hand, relations between sentences (\eg contradiction) sometimes have much simpler symmetries.
In the next section, we examine how global symmetries can be exploited in an inference task.

\section{Global Symmetries in Natural Language Inference}
The Stanford Natural Language Inference (SNLI, \citealp{snli2015}) dataset attempts to provide training and evaluation data for the task of categorising the logical relationship between a pair of sentences.
Systems must identify whether each hypothesis stands in a relation of \emph{entailment}, \emph{contradiction} or \emph{neutral} to its corresponding premise.
A number of neural net architectures have been proposed that effectively learn to make test set predictions based purely on patterns learned from the training data, without additional knowledge of the real world or of the logical structure of the task.
Here, we evaluate the Decomposable Attention Model (DAM, \citealp{dam2016}) in terms of its ability to extrapolate to novel instances, consisting of contradictions from the original test set which have been reversed.
For a human that understands the task, such generalisation is obvious: knowing that A contradicts B is equivalent to knowing that B contradicts A.
However, it is not at all clear that a model will learn this symmetry from the SNLI data, without it being imposed on the model in some way.
Consequently we also evaluate a modification, S-DAM, where this constraint is enforced by design.
\paragraph{Models.}
%

%
Both models build representations, $\mathbf{v}_p$ and $\mathbf{v}_h$, of premise and hypothesis in attend and compare steps.
The original DAM model then combines these representations by concatenating them 
and then transforming and aggregating the result to produce a final representation $\mathbf{u}_{ph}$, forming the input to a 3-way softmax:
\begin{equation}
 \begin{aligned}
   \mathbf{u}_{ph} & = t( \mathbf{v}_p ; \mathbf{v}_h ), \\
   p(i) & = s(\mathbf{u}_{ph} \cdot \mathbf{W}_i), \quad \text{with } i \in \{ c, e, n\}. \\
 \end{aligned}
\end{equation}

In S-DAM, we break the prediction into two decisions: contradiction vs. non-contradiction followed by entailment vs. neutral.
The first decision is symmetrised by concatenating the vectors in both orders and then summing the output of the same transformation applied to both concatenations:
\begin{equation}
 \begin{aligned}
   \tilde{\mathbf{u}}_{ph} & = t( \mathbf{v}_p ; \mathbf{v}_h ) + t( \mathbf{v}_h ; \mathbf{v}_p ), \\
   p(j) & = s(\tilde{\mathbf{u}}_{ph} \cdot \tilde{\mathbf{W}}_j), \quad \text{with } j \in \{ c, \neg c \}. \\
 \end{aligned}
\end{equation}
Predictions for entailment and neutral are then made conditioned on $\neg c$:
\begin{equation}
 \begin{aligned}
   \bar{\mathbf{u}}_{ph} & = t( \mathbf{v}_p ; \mathbf{v}_h ), \\
   p(k|\neg c) & = s(\bar{\mathbf{u}}_{ph} \cdot \bar{\mathbf{W}}_k), \quad \text{with } k \in \{ e, n \}. \\
 \end{aligned}
\end{equation}

\paragraph{Results.}
\begin{table}[t]
\centering
\begin{tabular}{lcc}
\toprule
 {\bf Instances} & {\bf DAM}  & {\bf S-DAM} \\
\midrule
 Whole Test Set & 86.71\% & 85.95\% \\ 
 Contradictions & 85.94\% & 85.69\%  \\
 Reversed Contradictions & 78.13\% & 85.20\% \\
\bottomrule
\end{tabular}
\caption{Accuracy on all instances, contradictions and reversed contradictions from the SNLI test set.}\label{snlitab}
\end{table}
\cref{snlitab} gives the accuracies for both models on the whole SNLI test set, the subset of contradictions, and the same set of contradictions reversed.
In the last row, the DAM model suffers a significant fall in performance when the contradictions are reversed. 
In comparison, the S-DAM's performance is almost identical on both sets.
%
%

Thus, the S-DAM model extrapolates more effectively because its architecture exploits a global symmetry of the relation between sentences in the task.
In the following section, we investigate a global symmetry within the representation of words.

\section{Global Structure in Word Embeddings}
Word embeddings, such as GloVe~\citep{glove2014} and word2vec~\citep{word2vec2013}, have been enormously effective as input representations for downstream tasks such as question answering or natural language inference. 
One well known application is the $king = queen - woman + man$ example, which represents an impressive extrapolation from word co-occurrence statistics to linguistic analogies~\citep{DBLP:conf/conll/LevyG14}.
To some extent, we can see this prediction as exploiting a global structure 
in which the differences between analogical pairs, such as $man-woman$, $king-queen$ and $father-mother$, are approximately equal.
Here, we consider how this global structure in the learned embeddings is 
related to a linearity in the training objective. 
In particular, linear functions have the property that $f(a+b) = f(a) + f(b)$, 
imposing a systematic relation between the predictions we make for $a$, $b$ and $a+b$. 
In fact, we could think of this as a form of translational symmetry 
where adding $a$ to the input has the same effect on the output throughout the space.

We hypothesise that breaking this linearity, and allowing a more local fit to the training data 
will undermine the global structure that the analogy predictions exploit.

\paragraph{Models.}
These embedding models typically rely on a simple dot product comparison of target and context vectors as the basis for predicting some measure of co-occurrence $s$:
\begin{equation}
    s = f \left( \sum_i \text{target}_i \cdot \text{context}_i\right).
\end{equation}
We replace this simple linear function of the context vectors, with a set of non-linear broken-stick functions $g_i({}\cdot{})$. 
\begin{equation*}
 \begin{aligned}
 s & = f \left( \sum_i g_i \left( \text{context}_i \right) \right), \\
 g_i \left( x \right) & =
  \begin{cases}
   m_i x & \text{if } n_i x + c_i < 0, \\
   \left(m_i + n_i \right) x + c_i & \text{otherwise.}
  \end{cases}
 \end{aligned}
\end{equation*}
We modify the CBOW algorithm in the publicly available word2vec code to incorporate this non-linearity and train on the commonly used \emph{text8} corpus of 17M words from Wikipedia.
As this modification doubles the number of parameters used for each word, we test models of dimensions 100, 200 and 400.

\paragraph{Results.}

\begin{table}[t]
\centering
\begin{tabular}{rll}
\toprule
 {\bf D} & {\bf Linear} & {\bf Non-Linear} \\
\midrule
 100 & 50.38\%           & 42.96\% \\
 200 & 53.18\%           & 40.66\% \\
 400 & 50.77\%           & 32.43\% \\
\bottomrule
\end{tabular}
\caption{Accuracy on the analogy task.}\label{msetab2}
\end{table}

\cref{msetab2} reports the performance on the standard analogy task distributed with the word2vec code. 
The non-linear modification of CBOW is substantially less successful than the original linear version on this task. 
This is true on all the sizes of models we evaluated, 
indicating that this decrease is not simply a result of over-parameterisation. 

Thus, destroying the global linearity in the embedding model undermines extrapolation to the analogy task.

\section{Conclusions}

Language is a very complex phenomenon, 
and many of its quirks and idioms need to be treated as local phenomena. 
However, we have also shown here examples in the representation of words and sentences where global structure supports extrapolation outside the training data. 

One tool for thinking about this dichotomy is the \emph{equivalent kernel} \cite{silverman1984}, which measures the extent to which a given prediction is influenced by nearby training examples. 
Typically, models with highly local equivalent kernels - e.g. splines, sigmoids and random forests - are preferred over non-local models - e.g. polynomials - in the context of general curve fitting \cite{hastieetal2001}. 

However, these latter functions are also typically those used to express fundamental scientific laws - e.g. $E=mc^2$, $F=G\frac{m_1 m_2}{r^2}$ - 
which frequently support extrapolation outside the original data from which they were derived. 
Local models, by their very nature, are less suited to making predictions outside the training manifold, as the influence of those training instances attenuates quickly.

We suggest that NLP will benefit from incorporating more global structure into its models. 
Existing background knowledge is one possible source for such additional structure \cite{marcus2018,minervinietal2017}. 
But it will also be necessary to uncover novel global relations, 
following the example of the other natural sciences.

We have used the development of the scientific understanding of planetary motion 
as a repeated example of the possibility of 
uncovering global structures that support extrapolation, throughout our discussion. 
Kepler and Newton found laws that went beyond simply maximising the fit 
to the known set of planetary bodies to describe regularities 
that held for every body, terrestrial and heavenly.

In our SNLI example, we showed that simply maximising the fit on the development and test sets does not yield a model that extrapolates to reversed contradictions.
In the case of word2vec, we showed that performance 
on the analogy task was related to the linearity 
in the objective function. 

More generally, we want to draw attention to the need for models in NLP that make meaningful predictions outside the space of the training data, and to argue that such extrapolation requires distinct modelling techniques from interpolation within the training space. 
Specifically, whereas the latter can often effectively rely on local smoothing between training instances, 
the former may require models that exploit global structures 
of the language phenomena.

\section*{Acknowledgments}

The authors are immensely grateful to Ivan Sanchez Carmona for many fruitful disagreements.
This work has been supported by the European Union H2020 project SUMMA (grant No. 688139), and by an Allen Distinguished Investigator Award.

\bibliographystyle{acl_natbib}
\bibliography{oddeven}

\begin{thebibliography}{19}
\expandafter\ifx\csname natexlab\endcsname\relax\def\natexlab#1{#1}\fi

\bibitem[{Abadi et~al.(2015)Abadi, Agarwal, Barham, Brevdo, Chen, Citro,
  Corrado, Davis, Dean, Devin, Ghemawat, Goodfellow, Harp, Irving, Isard, Jia,
  Jozefowicz, Kaiser, Kudlur, Levenberg, Man\'{e}, Monga, Moore, Murray, Olah,
  Schuster, Shlens, Steiner, Sutskever, Talwar, Tucker, Vanhoucke, Vasudevan,
  Vi\'{e}gas, Vinyals, Warden, Wattenberg, Wicke, Yu, and
  Zheng}]{tensorflow2015-whitepaper}
Mart\'{\i}n Abadi, Ashish Agarwal, Paul Barham, Eugene Brevdo, Zhifeng Chen,
  Craig Citro, Greg~S. Corrado, Andy Davis, Jeffrey Dean, Matthieu Devin,
  Sanjay Ghemawat, Ian Goodfellow, Andrew Harp, Geoffrey Irving, Michael Isard,
  Yangqing Jia, Rafal Jozefowicz, Lukasz Kaiser, Manjunath Kudlur, Josh
  Levenberg, Dandelion Man\'{e}, Rajat Monga, Sherry Moore, Derek Murray, Chris
  Olah, Mike Schuster, Jonathon Shlens, Benoit Steiner, Ilya Sutskever, Kunal
  Talwar, Paul Tucker, Vincent Vanhoucke, Vijay Vasudevan, Fernanda Vi\'{e}gas,
  Oriol Vinyals, Pete Warden, Martin Wattenberg, Martin Wicke, Yuan Yu, and
  Xiaoqiang Zheng. 2015.
\newblock {TensorFlow}: Large-scale machine learning on heterogeneous systems.
\newblock Software available from tensorflow.org.

\bibitem[{Bowman et~al.(2015)Bowman, Angeli, Potts, and Manning}]{snli2015}
Samuel~R. Bowman, Gabor Angeli, Christopher Potts, and Christopher~D. Manning.
  2015.
\newblock A large annotated corpus for learning natural language inference.
\newblock In \emph{Proceedings of the 2015 Conference on Empirical Methods in
  Natural Language Processing (EMNLP)}. Association for Computational
  Linguistics.

\bibitem[{Fodor and Pylyshyn(1988)}]{fodorpylyshyn1988}
Jerry~A. Fodor and Zenon~W. Pylyshyn. 1988.
\newblock Connectionism and cognitive architecture.
\newblock \emph{Cognition}, 28(1-2):3--71.

\bibitem[{Hastie et~al.(2001)Hastie, Tibshirani, and Friedman}]{hastieetal2001}
Trevor Hastie, Robert Tibshirani, and Jerome Friedman. 2001.
\newblock \emph{The Elements of Statistical Learning}.
\newblock Springer Series in Statistics. Springer New York Inc., New York, NY,
  USA.

\bibitem[{Hochreiter and Schmidhuber(1995)}]{flatopt1995}
Sepp Hochreiter and Jürgen Schmidhuber. 1995.
\newblock Simplifying neural nets by discovering flat minima.
\newblock In \emph{Advances in Neural Information Processing Systems 7}, pages
  529--536. MIT Press.

\bibitem[{Kiddon and Domingos(2015)}]{kiddondomingos2015}
Chlo{\'e} Kiddon and Pedro Rauel~C{\^a}ndido Domingos. 2015.
\newblock Symmetry-based semantic parsing.
\newblock Https://homes.cs.washington.edu/~pedrod/papers/sp14.pdf.

\bibitem[{{Lake} and {Baroni}(2017)}]{lakebaroni2017}
B.~M. {Lake} and M.~{Baroni}. 2017.
\newblock \href {http://arxiv.org/abs/1711.00350} {{Generalization without
  systematicity: \\ On the compositional skills of sequence-to-sequence
  recurrent networks}}.
\newblock \emph{ArXiv e-prints}.

\bibitem[{Lecun et~al.(1998)Lecun, Bottou, Bengio, and
  Haffner}]{Lecun98gradient-basedlearning}
Yann Lecun, Léon Bottou, Yoshua Bengio, and Patrick Haffner. 1998.
\newblock Gradient-based learning applied to document recognition.
\newblock In \emph{Proceedings of the IEEE}, pages 2278--2324.

\bibitem[{Levy and Goldberg(2014)}]{DBLP:conf/conll/LevyG14}
Omer Levy and Yoav Goldberg. 2014.
\newblock Linguistic regularities in sparse and explicit word representations.
\newblock In \emph{Proceedings of the Eighteenth Conference on Computational
  Natural Language Learning, CoNLL 2014}, pages 171--180.

\bibitem[{{Marcus}(2018{\natexlab{a}})}]{marcus2017}
G.~{Marcus}. 2018{\natexlab{a}}.
\newblock \href {http://arxiv.org/abs/1801.00631} {{Deep Learning: A Critical
  Appraisal}}.
\newblock \emph{ArXiv e-prints}.

\bibitem[{{Marcus}(2018{\natexlab{b}})}]{marcus2018}
G.~{Marcus}. 2018{\natexlab{b}}.
\newblock \href {http://arxiv.org/abs/1801.05667} {{Innateness, AlphaZero, and
  Artificial Intelligence}}.
\newblock \emph{ArXiv e-prints}.

\bibitem[{Marcus(1998)}]{marcus1998}
Gary~F. Marcus. 1998.
\newblock Rethinking eliminative connectionism.
\newblock \emph{Cognitive Psychology}, 37:243--282.

\bibitem[{Mikolov et~al.(2013)Mikolov, Sutskever, Chen, Corrado, and
  Dean}]{word2vec2013}
Tomas Mikolov, Ilya Sutskever, Kai Chen, Greg Corrado, and Jeffrey Dean. 2013.
\newblock Distributed representations of words and phrases and their
  compositionality.
\newblock In \emph{Proceedings of the 26th International Conference on Neural
  Information Processing Systems - Volume 2}, NIPS'13, pages 3111--3119, USA.
  Curran Associates Inc.

\bibitem[{Minervini et~al.(2017)Minervini, Demeester, Rockt{\"{a}}schel, and
  Riedel}]{minervinietal2017}
Pasquale Minervini, Thomas Demeester, Tim Rockt{\"{a}}schel, and Sebastian
  Riedel. 2017.
\newblock \href {http://www.auai.org/uai2017/media/Booklet.pdf} {Adversarial
  sets for regularising neural link predictors}.
\newblock In \emph{Proceedings of the Thirty-Third Conference on Uncertainty in
  Artificial Intelligence, {UAI} 2017, Sydney, Australia, August 11-15, 2017}.
  {AUAI} Press.

\bibitem[{Parikh et~al.(2016)Parikh, T{\"{a}}ckstr{\"{o}}m, Das, and
  Uszkoreit}]{dam2016}
Ankur~P. Parikh, Oscar T{\"{a}}ckstr{\"{o}}m, Dipanjan Das, and Jakob
  Uszkoreit. 2016.
\newblock A decomposable attention model for natural language inference.
\newblock In \emph{Proceedings of the 2016 Conference on Empirical Methods in
  Natural Language Processing, {EMNLP} 2016, Austin, Texas, USA, November 1-4,
  2016}, pages 2249--2255.

\bibitem[{Pennington et~al.(2014)Pennington, Socher, and Manning}]{glove2014}
Jeffrey Pennington, Richard Socher, and Christopher~D. Manning. 2014.
\newblock Glove: Global vectors for word representation.
\newblock In \emph{In EMNLP}.

\bibitem[{Silverman(1984)}]{silverman1984}
B.~W. Silverman. 1984.
\newblock \href {https://doi.org/10.1214/aos/1176346710} {Spline smoothing: The
  equivalent variable kernel method}.
\newblock \emph{Ann. Statist.}, 12(3):898--916.

\bibitem[{Srivastava et~al.(2014)Srivastava, Hinton, Krizhevsky, Sutskever, and
  Salakhutdinov}]{dropout2014}
Nitish Srivastava, Geoffrey Hinton, Alex Krizhevsky, Ilya Sutskever, and Ruslan
  Salakhutdinov. 2014.
\newblock Dropout: A simple way to prevent neural networks from overfitting.
\newblock \emph{Journal of Machine Learning Research}, 15:1929--1958.

\bibitem[{Tikhonov(1963)}]{l2reg1963}
A.~N. Tikhonov. 1963.
\newblock Solution of incorrectly formulated problems and the regularization
  method.
\newblock \emph{Soviet Math. Dokl.}, 4:1035--1038.

\end{thebibliography}

\end{document}